\documentclass[sigplan,10pt]{acmart}
\acmConference[PPoPP'26]{ACM SIGPLAN Annual Symposium on Principles and Practice of Parallel Programming}{January 31--February 04, 2026}{Sydney, Australia}
\acmYear{2026}
\acmISBN{} 
\acmDOI{} 
\startPage{1}

\setcopyright{none}

\usepackage{booktabs}   
\usepackage{subcaption} 

\usepackage{amsmath}
\usepackage{algorithm}
\usepackage[noend]{algpseudocode}
\usepackage{listings}
\usepackage{xcolor}

\definecolor{codegreen}{rgb}{0,0.6,0}
\definecolor{codegray}{rgb}{0.5,0.5,0.5}
\definecolor{codepurple}{rgb}{0.58,0,0.82}
\definecolor{backcolour}{rgb}{0.95,0.95,0.92}

\lstdefinestyle{mystyle}{
    backgroundcolor=\color{backcolour},   
    commentstyle=\color{codegreen},
    keywordstyle=\color{magenta},
    numberstyle=\tiny\color{codegray},
    stringstyle=\color{codepurple},
    basicstyle=\ttfamily\footnotesize,
    breakatwhitespace=false,         
    breaklines=true,                 
    captionpos=b,                    
    keepspaces=true,                 
    numbers=left,                    
    numbersep=5pt,                  
    showspaces=false,                
    showstringspaces=false,
    showtabs=false,                  
    tabsize=2
}
\begin{document}

\title[BuddyMoE]{BuddyMoE: Exploiting Expert Redundancy to Accelerate Memory-Constrained Mixture-of-Experts Inference}         


\author{
    \begin{tabular}{cccc} 
        Yun Wang & Lingyun Yang & Senhao Yu & Yixiao Wang \\
        Ruixing Li & Zhixiang Wei & James Yen & Zhengwei Qi \\
    \end{tabular}
}
\affiliation{
   \institution{Shanghai Jiao Tong University}
   \city{Shanghai}
   \country{China}
}
\fancyhead{}  
\renewcommand\footnotetextcopyrightpermission[1]{} 

\begin{abstract}

Mixture-of-Experts (MoE) architectures scale language models by activating only a subset of specialized expert networks for each input token, thereby reducing the number of floating-point operations. However, the growing size of modern MoE models causes their full parameter sets to exceed GPU memory capacity; for example, Mixtral-8x7B has 45 billion parameters and requires 87 GB of memory even though only 14 billion parameters are used per token. Existing systems alleviate this limitation by offloading inactive experts to CPU memory, but transferring experts across the PCIe interconnect incurs significant latency ($\sim$10 ms). Prefetching heuristics aim to hide this latency by predicting which experts are needed, but prefetch failures introduce significant stalls and amplify inference latency. In the event of a prefetch failure, prior work offers two primary solutions: either fetch the expert on demand, which incurs a long stall due to the PCIe bottleneck, or drop the expert from the computation, which significantly degrades model accuracy. The critical challenge, therefore, is to maintain both high inference speed and model accuracy when prefetching fails.

In this paper, we propose \emph{BuddyMoE}, a system that harnesses expert redundancy to mitigate prefetch‑miss penalties during MoE inference. BuddyMoE identifies pairs of \textit{buddy experts} that have similar functionality by analyzing co‑activation patterns across tokens.  When prefetching mispredicts and an expert would be loaded on demand, the system dynamically substitutes a cached buddy expert, trading a small accuracy loss for substantial throughput gains. We present algorithms for finding similar experts, analyze the resulting accuracy degradation, and design a runtime replacement mechanism that balances model accuracy against inference latency. Experiments on state‑of‑the‑art MoE models demonstrate that BuddyMoE reduces expert miss latency and improves up to 10\% throughput (tokens‑per‑second) with negligible accuracy loss.

\end{abstract}

\maketitle

\section{Introduction}

Modern large language models (LLMs) have achieved remarkable capabilities by scaling to hundreds of billions of parameters~\cite{zhong2025flashtensor,frantar2025marlin}. However, this scaling presents a significant computational challenge, as training and inferencing such dense models are beyond the reach of most organizations~\cite{wang2025ccfuser,lin2025weipipe}. The Mixture-of-Experts (MoE) architecture has emerged as a compelling solution by decoupling model capacity from computational cost~\cite{shazeer2017outrageously,lepikhin2021gshard,fedus2021switch}. MoE architectures compose sparse layers from a large pool of smaller "expert" networks. For each input token, a gating network dynamically activates only a small subset of these experts, dramatically reducing the floating-point operations (FLOPs) required. This approach has proven effective, underpinning widely-deployed models like Mixtral, QWEN3, and DeepSeek-MoE~\cite{lin2025weipipe,frantar2025marlin,DBLP:journals/corr/abs-2412-19437,DBLP:journals/corr/abs-2505-09388}.

\begin{figure}[t!]
    \centering
    \includegraphics[width=1.0\linewidth]{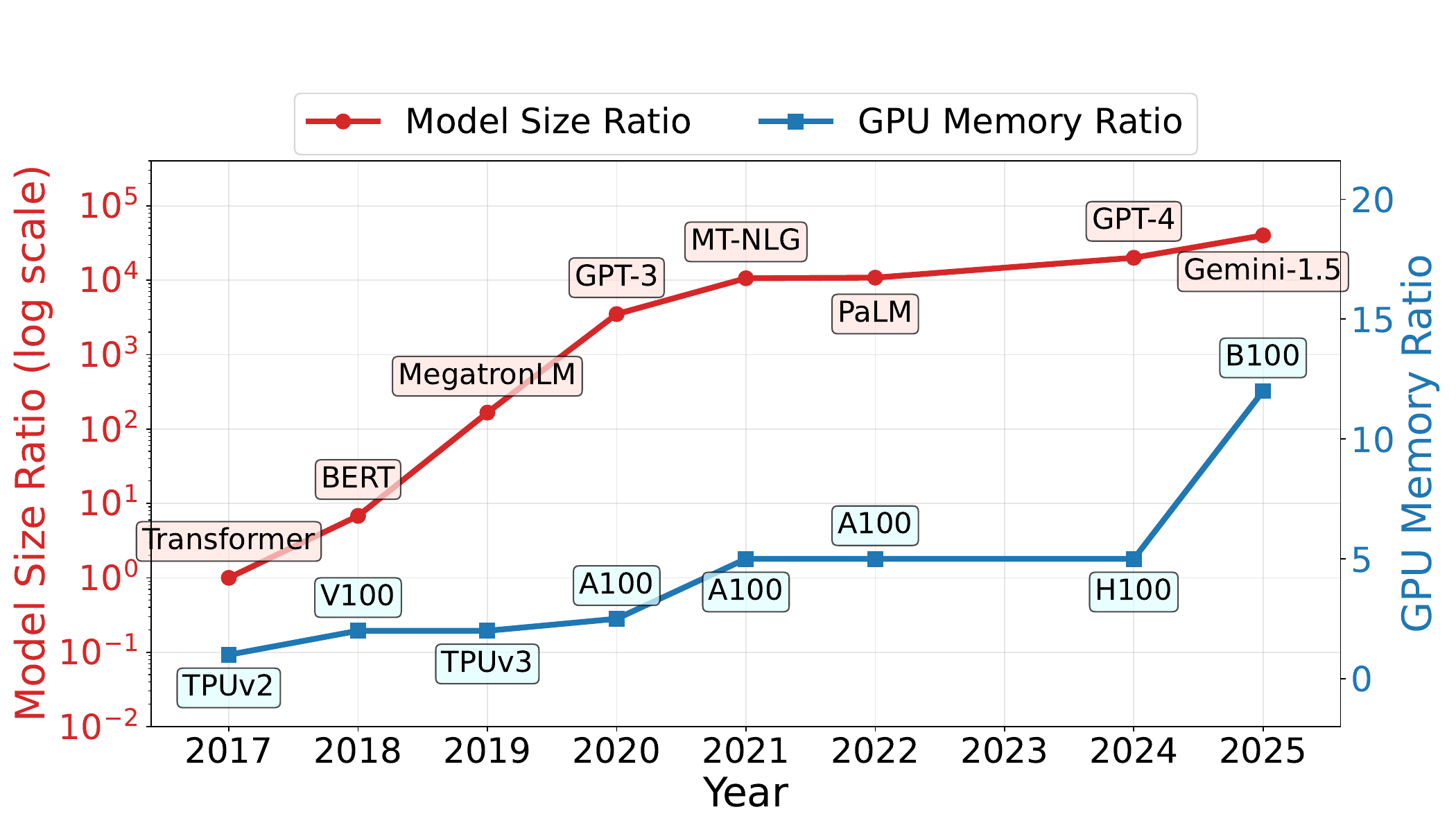}
    \caption{Model size is scaling substantially faster than single-accelerator memory (2017–2025). Left (log) axis: relative model size; right axis: relative device memory, illustrating the widening gap.}
    \label{fig:llm_scaling}
    \vspace{-1em}
\end{figure}

Despite their training efficiency, MoE models introduce a critical inference challenge: memory inefficiency. Because the gating network can route a token to \emph{any} expert, the parameters for all experts must reside in high-bandwidth but limited GPU memory~\cite{zhong2025flashtensor,zhang2025sgdrc}. This is true even though only a fraction of experts are active for any given token. As models incorporate more experts, their total parameter count often exceeds single-GPU memory capacity—a problem exacerbated by the widening gap between model size and device memory shown in Figure~\ref{fig:llm_scaling}. For instance, Mixtral-8×7B activates only 14 billion parameters per forward pass, yet its full 45 billion parameters require approximately 87 GB of memory for deployment.

To deploy these large models on memory-constrained hardware, systems like Zero-Infinity and Accelerate have pioneered parameter offloading, relocating inactive experts to slower, more abundant storage tiers like CPU memory~\cite{rajbhandari2021zero,hf_accelerate_infer_2025,hf_accelerate_deepspeed_2025}. While this reduces GPU memory pressure, it introduces a severe performance bottleneck: data transfer latency~\cite{liu2025mario,sun2025compso}. Loading an expert from CPU memory via PCIe can take tens of milliseconds, whereas the corresponding GPU computation takes only a few. This disparity transforms inference from a compute-bound task to an I/O-bound one, fundamentally limiting throughput.

Recent systems attempt to mitigate this I/O bottleneck with predictive prefetching, which anticipates upcoming expert needs and loads them in the background. However, these strategies face fundamental limitations~\cite{liu2025doradd}. Expert selection is context-dependent and difficult to predict accurately. A misprediction forces a synchronous load, incurring the full I/O latency, and potentially negating the benefits of many successful prefetches. 

Our approach stems from a crucial empirical observation: experts within large MoE models exhibit substantial functional redundancy. This redundancy, confirmed by prior work showing that MoE models tolerate aggressive pruning (down to 4bits) with minimal quality loss~\cite{frantar2025marlin}, means multiple experts often learn similar functions. This presents an unexplored opportunity: instead of stalling to fetch a missing expert, the system could substitute a functionally similar expert that is already resident in GPU memory. To our knowledge, \texttt{BuddyMoE} is the first to utilize this redundancy to gracefully mitigate prefetch failures.

Expert prefetch misses in large inference models create significant performance bottlenecks. To address this, we present \textbf{BuddyMoE}, a novel system that exploits expert redundancy to replace these slow operations with efficient approximations. We formalize this redundancy through the concept of \emph{buddy experts}: functionally similar experts that are identified via an offline analysis of co-activation patterns and output similarity on a calibration dataset. During inference, BuddyMoE uses a pre-computed lookup table to intercept a failed prefetch. Rather than waiting for a synchronous load from slower memory, it instantly substitutes the required expert with a buddy that is already available on the GPU. This approach yields a substantial reduction in latency for a marginal and often imperceptible impact on model accuracy. This paper details our methods for robust buddy identification and systematically characterizes the resulting performance-accuracy trade-off~\cite{liang2025attnchecker}.

Our work makes the following contributions:
\begin{itemize}
    \item We introduce the novel concept of \emph{buddy experts} and propose a methodology for identifying functionally similar experts using co-activation frequency and output similarity metrics.
    \item We conduct a comprehensive analysis of the impact of buddy expert substitution on model accuracy, characterizing how this trade-off is influenced by the number of replacements and the sensitivity of different tokens.
    \item We design and implement BuddyMoE, a runtime system that integrates buddy substitution into the inference pipeline to mitigate prefetch-miss penalties. Our experiments on state-of-the-art MoE models demonstrate that BuddyMoE significantly reduces miss-related latency and improves up to 10\% throughput with minimal loss in model quality.
\end{itemize}

\section{Background and Motivation}

\begin{figure}[t!]
    \centering
    \begin{subfigure}[b]{0.85\linewidth} 
        \includegraphics[width=\textwidth]{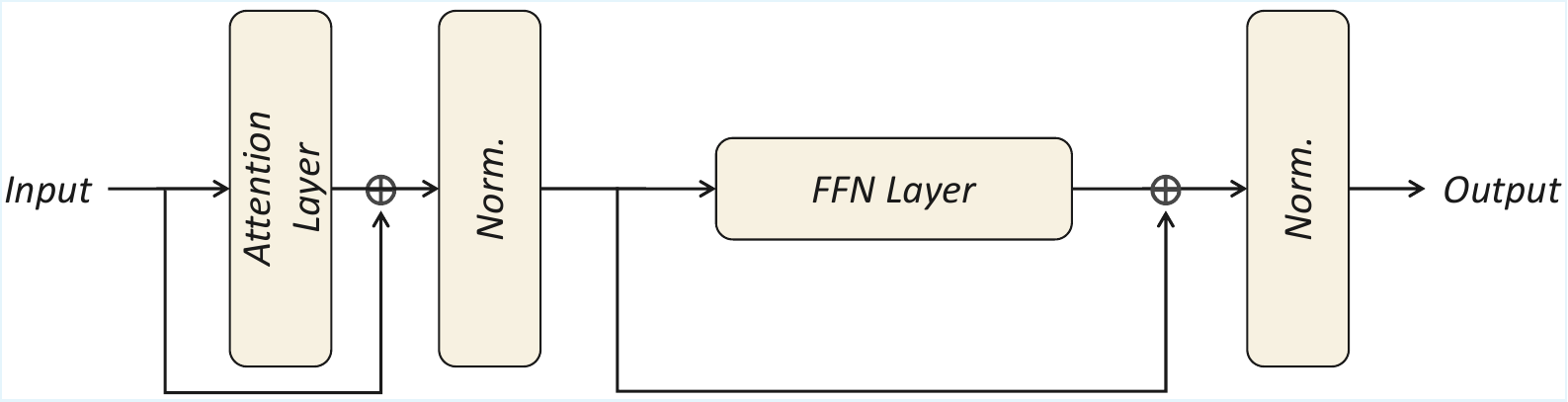} 
        \caption{A standard Transformer block with a dense Feed-Forward Network (FFN).}
        \label{fig:dense_transformer}
    \end{subfigure}
    
    \vspace{1em} 
    
    \begin{subfigure}[b]{0.85\linewidth} 
        \includegraphics[width=\textwidth]{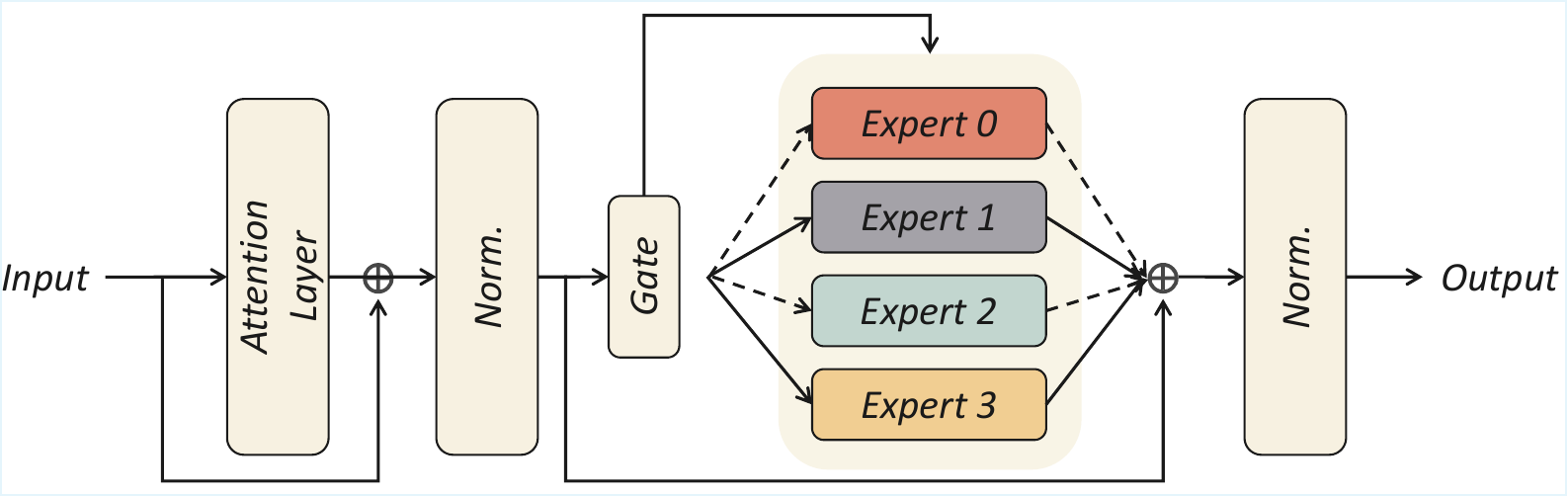} 
        \caption{An MoE block where the dense FFN is replaced by a pool of sparsely activated experts controlled by a gating network.}
        \label{fig:moe_transformer}
    \end{subfigure}
    
    \caption{Architectural comparison between a standard Transformer block (a) and a MoE block (b). The MoE architecture replaces the single, dense FFN with a pool of experts, only a subset of which are activated per token.}
    \label{fig:transformer_vs_moe}
\end{figure}

\subsection{Transformers and Mixture‑of‑Experts}

Transformer architecture, the basis for modern LLMs, utilizes self-attention and dense feed-forward networks (FFNs). In a standard dense Transformer block (Figure~\ref{fig:dense_transformer}), every input token is processed by a single FFN, activating all its parameters. Scaling these FFNs enhances performance, but their dense computation becomes very costly as models get larger.
MoE offers a more efficient scaling method~\cite{shazeer2017outrageously,lepikhin2021gshard,fedus2021switch}. An MoE Transformer (Figure~\ref{fig:moe_transformer}) replaces the dense FFN with an MoE layer. This layer has a pool of independent expert networks (FFNs) and a small gating network. The gating network dynamically selects a small number of experts—usually one or two—for each input token. This sparse activation allows for a massive increase in total parameters without a proportional increase in computational cost per token. As a result, MoE models can match the performance of much larger dense models with less inference computation, though this architecture creates a major memory footprint problem for deployment~\cite{wang2025ccfuser,lin2025weipipe}.

\subsection{Mixture‑of‑Experts and Offloading}

\begin{figure}[t!]
    \centering
    \includegraphics[width=1.0\linewidth]{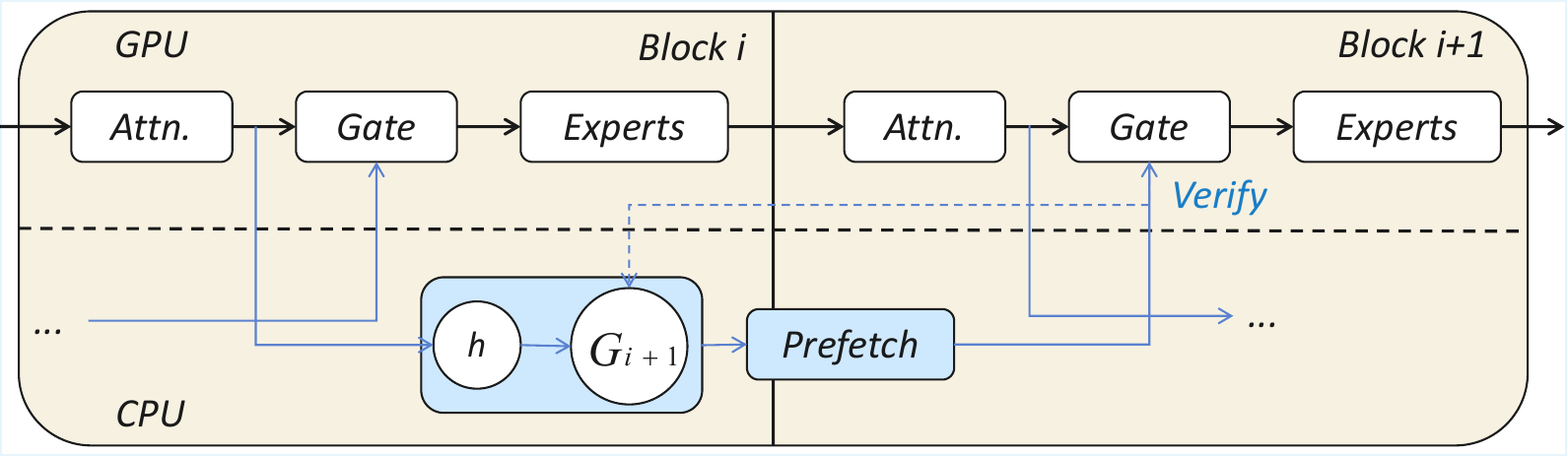}
    \caption{Overview of the expert prefetching pipeline. While the GPU computes block i, the CPU uses the attention output to predict the required experts for the next block (i+1) and prefetches them. This overlaps I/O with computation to hide latency. A verification step is used to handle prediction mismatches.}
    \label{fig:expert_prefetching}
    \vspace{-1em}
\end{figure}

\begin{table}[t]
\centering
\caption{Impact of Cache Misses and BuddyMoE on MoE Inference. BuddyMoE's buddy replacement strategy mitigates the latency penalty of a prefetch miss.}
\label{tab:cache_penalty_compact}
\small
\setlength{\tabcolsep}{6pt}
\begin{tabular}{lcc}
\toprule
\textbf{Scenario} & \textbf{Latency (ms)} & \textbf{Accuracy} \\
\midrule
Baseline (On Demand) & 9-10 & Lossless \\
Prefetch Hit & $\sim$0 & Lossless \\
Prefetch Miss & 9-10 & Lossless \\
BuddyMoE Hit & $\sim$0 & Lossless \\
BuddyMoE Miss & $\sim$0 & Minimal Loss \\
\bottomrule
\end{tabular}
\end{table}



MoE layers consist of a gating network and multiple experts. For each token, the gate assigns probabilities and activates the top-$k$ experts, whose outputs are aggregated. This sparse activation saves computation compared to dense feed-forward networks, but all expert parameters must still reside in GPU memory for possible activation, creating heavy memory demands~\cite{zhong2025flashtensor}. Large MoE models like Mixtral-8×7B have billions of parameters; though only a few experts run per token, the full model still requires 87 GB.

To fit such models on GPUs with limited memory, offloading places inactive experts in CPU memory or storage. The GPU holds frequently used experts (the \emph{expert cache}) and non-expert weights, fetching others over PCIe when needed. While this reduces GPU usage, it adds large latency: transferring one expert from CPU takes tens of milliseconds, far exceeding a single MoE layer’s compute time~\cite{liu2025mario,zhang2025sgdrc}.

\subsection{Expert Prefetching}
To reduce pipeline stalls from on‑demand loading, prefetching systems attempt to anticipate expert requirements and transfer them to the GPU cache ahead of time.  Techniques employed by systems like MoE‑Infinity and Pre‑gated MoE use signals such as historical activation frequencies or auxiliary gating logic to inform these predictions~\cite{xue2024moe,hwang2024pregatedmoe}.  When successful, prefetching effectively hides I/O latency by overlapping it with computation.  However, the input‑dependent nature of expert routing makes perfect prediction infeasible, leading to two primary sources of inefficiency.  First, a \emph{prediction miss} occurs when a required expert is not in the cache, forcing a synchronous fetch that stalls execution.  Second, \emph{speculative waste} arises from loading experts that are not ultimately used, consuming finite PCIe bandwidth and displacing other potentially useful experts~\cite{sun2025compso,liu2025doradd}.

\subsection{Motivation}

\begin{figure}[t!]
    \centering
    \includegraphics[width=1.0\linewidth]{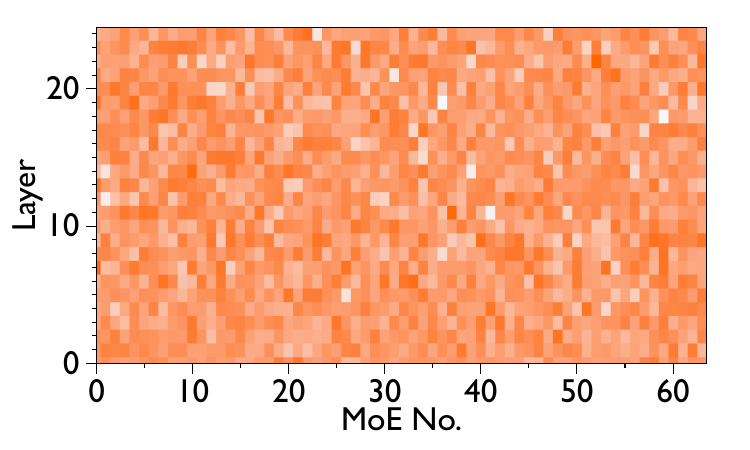}
    \caption{Expert similarity heatmap for a 64-expert MoE model. The intensity represents the functional similarity between expert pairs, with brighter regions indicating higher similarity. The prevalent bright areas demonstrate significant redundancy across experts, suggesting opportunities for expert substitution during cache misses.}
    \label{fig:expert_similarity}
    \vspace{-1em}
\end{figure}

The efficiency of MoE inference under memory constraints depends critically on balancing limited GPU memory against the latency penalty of expert loading. Our work is motivated by three key observations that together suggest a new approach to this challenge.

\noindent\textbf{Inherent expert redundancy.} Figure~\ref{fig:expert_similarity} shows a similarity analysis of experts in a production MoE model, revealing substantial functional overlap between expert pairs. The bright regions in the heatmap highlight many experts with high similarity scores, indicating that the model’s capacity is redundantly distributed. Prior work confirms that aggressive quantization or pruning of selected experts causes minimal accuracy loss, further validating this redundancy~\cite{frantar2025marlin}. This suggests that experts with overlapping functionality could substitute for one another during inference, especially when memory constraints prevent loading all experts.

\noindent\textbf{Disproportionate cost of cache misses.} Profiling MoE inference shows that expert loading dominates the pipeline. On edge devices running Mixtral-8×7B, transfers from CPU memory account for 85–94\% of inference latency, while computation by the experts themselves is negligible~\cite{liu2025mario,zhang2025sgdrc}. This imbalance stems from PCIe bandwidth limits (16–32 GB/s versus TB/s within GPU memory). A prefetch miss stalls the pipeline for tens of milliseconds while waiting for the transfer, delaying the entire inference process. Prefetching systems attempt to mask this cost, but their effectiveness is bounded by the stochasticity of expert routing and imperfect predictions.

\noindent\textbf{Limitations of predictive prefetching.} Prefetching can overlap I/O with compute when predictions are accurate, but MoE’s input-dependent routing makes perfect prediction unattainable. Current heuristics—ranging from activation frequency tracking to auxiliary gating networks—suffer from two core inefficiencies: prediction misses force synchronous expert loads that stall the pipeline, while speculative prefetching wastes PCIe bandwidth and may evict more useful experts from cache~\cite{sun2025compso,liu2025doradd}. These problems intensify as the number of experts grows and activation patterns become harder to anticipate.

Together, these observations motivate BuddyMoE’s core insight: instead of eliminating cache misses via better prediction, we can tolerate them gracefully by exploiting expert redundancy. By caching functionally similar “buddy” experts, we can substitute a cached buddy when the target expert is unavailable, avoiding catastrophic stalls while preserving acceptable accuracy. In this design, cache misses degrade into minor perturbations absorbed by the model’s inherent redundancy.

\noindent\textbf{Challenges.} Exploiting expert redundancy for miss tolerance introduces two challenges. First, we must rigorously define “buddy” similarity and identify expert pairs beyond simple heuristics such as output variance or access frequency. Second, substitution unavoidably perturbs outputs, so we need mechanisms to bound accuracy loss while maximizing latency reduction. Addressing this trade-off is central to realizing BuddyMoE’s potential~\cite{liang2025attnchecker}.

\begin{figure}[t!]
    \centering
    \includegraphics[width=1.0\linewidth]{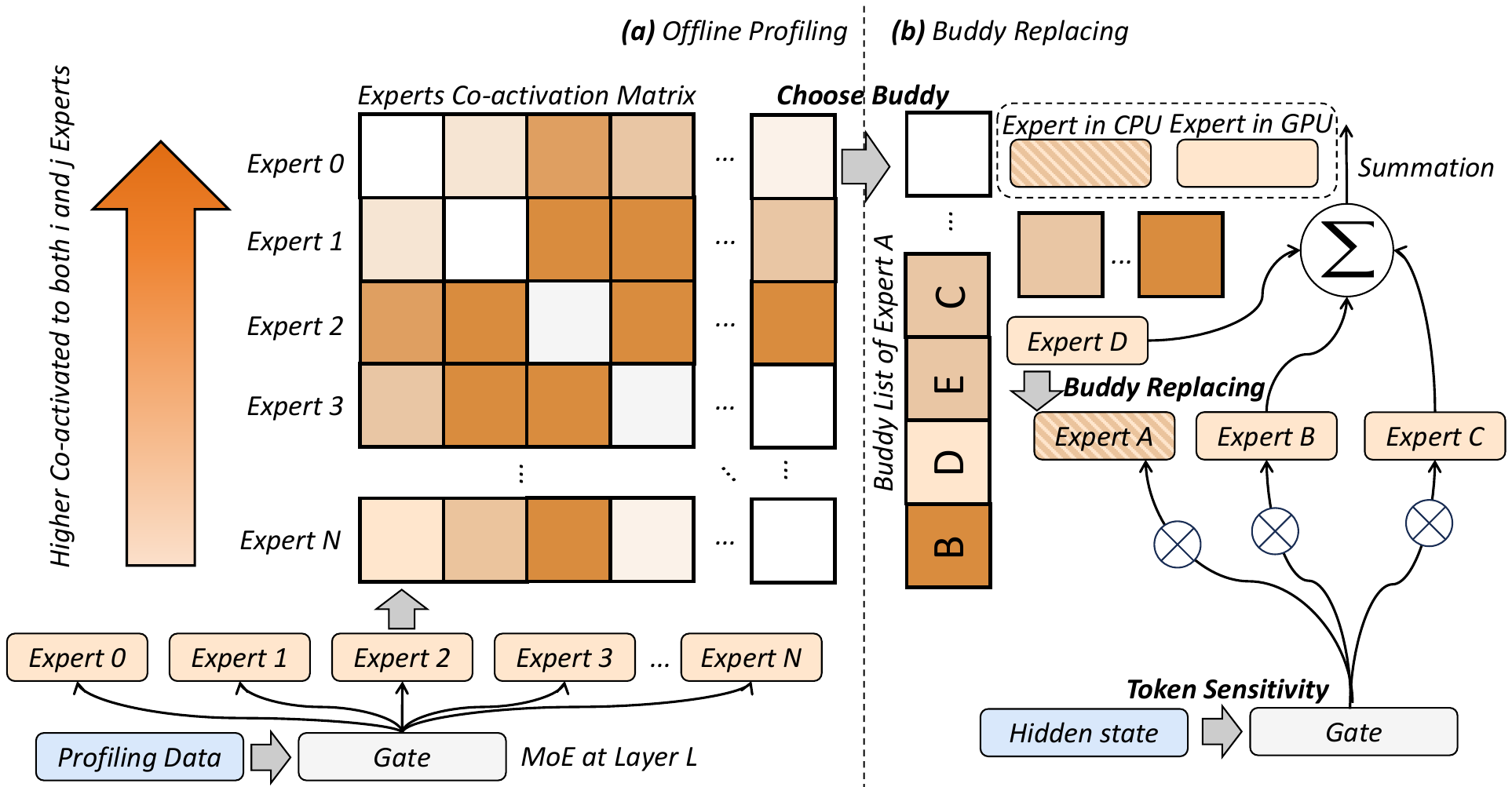}
    \caption{Overview of the buddy expert replacement system. The left panel shows the experts co-activation matrix derived from profiling data, where darker cells indicate higher co-activation frequency between expert pairs. The right panel illustrates the buddy replacing mechanism: (a) offline profiling identifies frequently co-activated experts with functional redundancy, and (b) runtime buddy replacing selectively substitutes GPU-resident experts with their CPU-based buddies based on activation patterns, token sensitivity, and expert distribution. The system dynamically determines replacement decisions through three key metrics (§3.1-§3.3) to minimize memory transfer overhead while maintaining model accuracy.}
    \label{fig:expert_prefetching}
    \vspace{-1em}
\end{figure}

\section{System Design}

BuddyMoE is a replacement-based runtime for MoE inference. It operationalizes two robust empirical regularities of MoE routers: (i) \textbf{functional redundancy}---multiple experts implement similar behaviors on overlapping token sub-manifolds; and (ii) \textbf{uneven activation and co-activation}---a small fraction of experts (and expert pairs) accounts for most routing events. BuddyMoE converts these properties into throughput gains under memory pressure by replacing missing or CPU-resident experts with \emph{buddy} experts that are already on the GPU, subject to accuracy-preserving gates.

\paragraph{Notation and Preliminaries.}
Let $\mathcal{L}$ be the MoE layers and $\mathcal{E}_\ell=\{1,\dots,E_\ell\}$ the experts at layer $\ell\in\mathcal{L}$. For token $x$, the router at layer $\ell$ emits logits $\mathbf{z}_\ell(x)\in\mathbb{R}^{E_\ell}$ and probabilities $\mathbf{p}_\ell(x)=\mathrm{softmax}(\mathbf{z}_\ell(x))$. With top-$k$ routing, the selected set is $S_\ell(x)\subset\mathcal{E}_\ell$, $|S_\ell(x)|=k$, and $\tilde{\mathbf{p}}_\ell(x)$ denotes $\mathbf{p}_\ell(x)$ renormalized over $S_\ell(x)$. The memory residency of expert $i$ is $\mathrm{loc}_\ell(i)\in\{\textsc{gpu},\textsc{cpu}\}$. For micro-batch $\mathcal{B}$, the set of experts requested at layer $\ell$ is $\mathcal{R}_\ell(\mathcal{B})$.

\subsection{Three Key Metrics for Replacement Decisions}
\label{sec:three-metrics}

BuddyMoE employs three key metrics executed in a strict sequence to determine whether and how to replace CPU-resident experts with their GPU-resident buddies. These metrics balance model accuracy preservation with memory transfer minimization.

\paragraph{Token Activating Entropy (TAE, $\tau$).}
The first metric quantifies token-level tolerance to expert substitution through Token Activating Entropy:
\begin{equation}
\mathrm{TAE}_\ell(x) \;=\; 
-\sum_{i\in S_\ell(x)} \tilde{p}_\ell(i\mid x)\,\log \tilde{p}_\ell(i\mid x) \Big/ \log k
\;\in [0,1].
\label{eq:tae}
\end{equation}
Low TAE indicates peaky routing where the token has strong preference for specific experts and is thus \emph{sensitive} to replacement. High TAE indicates diffuse routing with several comparable experts, making the token \emph{tolerant} to substitution. We forbid replacement for token $x$ at layer $\ell$ when $\mathrm{TAE}_\ell(x)\le\tau$ and allow it otherwise. Implementation details include: (i) using renormalized top-$k$ probabilities $\tilde{\mathbf{p}}_\ell(x)$ to avoid artifacts from the tail; (ii) optional temperature smoothing with $\tilde{\mathbf{p}}_\ell(x;T)=\mathrm{softmax}(\mathbf{z}_\ell(x)/T)$ restricted to $S_\ell(x)$ to stabilize TAE across layers ($T\!\in\![0.8,1.2]$ works well); (iii) percentile calibration where $\tau$ is picked as the $p$-th percentile of the per-layer TAE distribution (for $p\!\in\![10,20]$), making the gate robust across models and domains. For deployments requiring extra caution, TAE can be combined with a probability margin $m_\ell(x)=\tilde{p}_{\max}-\tilde{p}_{\text{2nd}}$ to forbid replacement if $(\mathrm{TAE}\le\tau)\ \lor\ (m_\ell(x)\ge\gamma)$.

\paragraph{Expert Distribution Gate (CPU/GPU Residency, $\beta$).}
The second metric provides batch-aware residency signaling to prevent broad, risky replacements. For micro-batch $\mathcal{B}$ at layer $\ell$, we define the fraction of currently required experts that are on CPU:
\begin{equation}
\delta_\ell(\mathcal{B}) \;=\;
\frac{\bigl|\{i\in \mathcal{R}_\ell(\mathcal{B}) \,:\, \mathrm{loc}_\ell(i)=\textsc{cpu}\}\bigr|}
     {\bigl|\mathcal{R}_\ell(\mathcal{B})\bigr|}\,.
\label{eq:delta}
\end{equation}
Given threshold $\beta\in[0,1]$, we bypass replacement when $\delta_\ell(\mathcal{B})\ge\beta$ and proceed otherwise. The intuition is that when many requested experts are on CPU, broad replacement would affect many tokens simultaneously and risks compounding errors. When only a few are on CPU, targeted replacement avoids bursty CPU $\!\to\!$ GPU traffic with limited accuracy exposure. The $\beta$ threshold can be related to a transfer budget: let $B_{\text{PCIe}}$ be an allowed per-step transfer budget and $\bar{w}$ the average bytes per expert (weights plus optimizer state if any). Let $\hat{n}_\text{cpu}(\ell)$ estimate the number of CPU-only invocations without replacement. Choosing $\beta$ that maintains $\hat{n}_\text{cpu}(\ell)\cdot \bar{w}\le B_{\text{PCIe}}$ keeps the system within bandwidth budget. In practice, a fixed $\beta$ per deployment tier suffices, though adaptive $\beta$ based on a running bandwidth meter is a drop-in extension. In tensor/pipeline parallel setups, we compute $\delta_\ell$ per partition and apply the gate locally.

\paragraph{Buddy Selection Priority Score ($\Psi$).}
If the token passes the TAE gate and the batch passes the distribution gate, and a requested expert $i\in S_\ell(x)$ is not on GPU, we attempt replacement with a buddy $j\in \mathcal{B}_\ell(i;\alpha)$ that is currently GPU-resident. Candidates are prioritized using:
\begin{equation}
\Psi_\ell(j\,|\,i,x) = \underbrace{q_{j\mid i}}_{\text{Global Sim.}} \!\cdot\!
\underbrace{\bigl(1+\eta\,\hat{z}_\ell^{(j)}(x)\bigr)}_{\text{Local Compat.}} \!\cdot\!
\underbrace{\bigl(1-\kappa\,\mathrm{hop}(j)\bigr)}_{\substack{\text{Topology/}\\\text{Cross-link Pen.}}},
\label{eq:psi}
\end{equation}
where $\hat{z}_\ell^{(j)}(x)$ is a normalized router logit for $j$ on token $x$ (if available), $\mathrm{hop}(j)$ counts cross-partition hops (0 for same-GPU), and $\eta,\kappa\!\ge\!0$ are small tunables (default $\eta=\kappa=0$). For top-$k$ routing, multiple missing experts from $S_\ell(x)$ might map to the same buddy. To avoid collapsing diversity, we discourage reusing an already-chosen buddy for that token by reducing its $\Psi_\ell$ multiplicatively (e.g., by a factor $<1$) on subsequent picks for the same $x$. If no GPU-resident buddy exists, we fall back to either prefetching the original $i$ (paying transfer cost) or skipping the expert per the baseline MoE drop policy, depending on SLA and the router's token capacity settings.

\subsection{Expert Co-activation Matrix Analysis}
\label{sec:coactivation-matrix}

\begin{figure}[t!]
    \centering
    \includegraphics[width=1.0\linewidth]{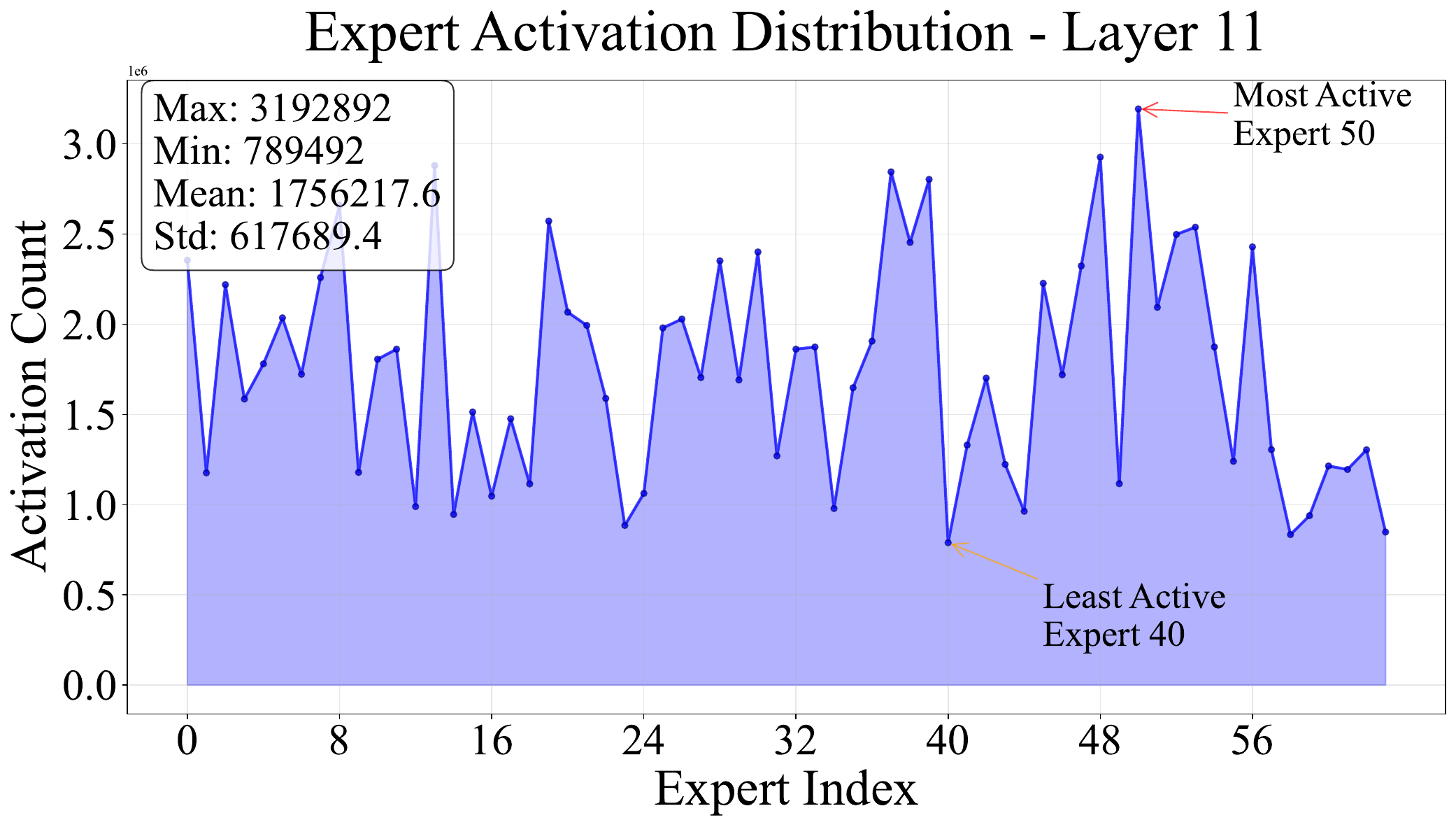}
    \caption{Uneven expert activation distribution in Layer 11 of a 64-expert MoE model. The activation count varies significantly across experts, with a few "popular" experts (e.g., Expert 50) accounting for a disproportionately large share of routing events compared to less active ones (e.g., Expert 40). This skew is a key empirical regularity exploited by BuddyMoE.}
    \label{fig:expert_activation_distribution}
    \vspace{-1em}
\end{figure}

\begin{figure}[t!]
    \centering
    \includegraphics[width=0.9\linewidth]{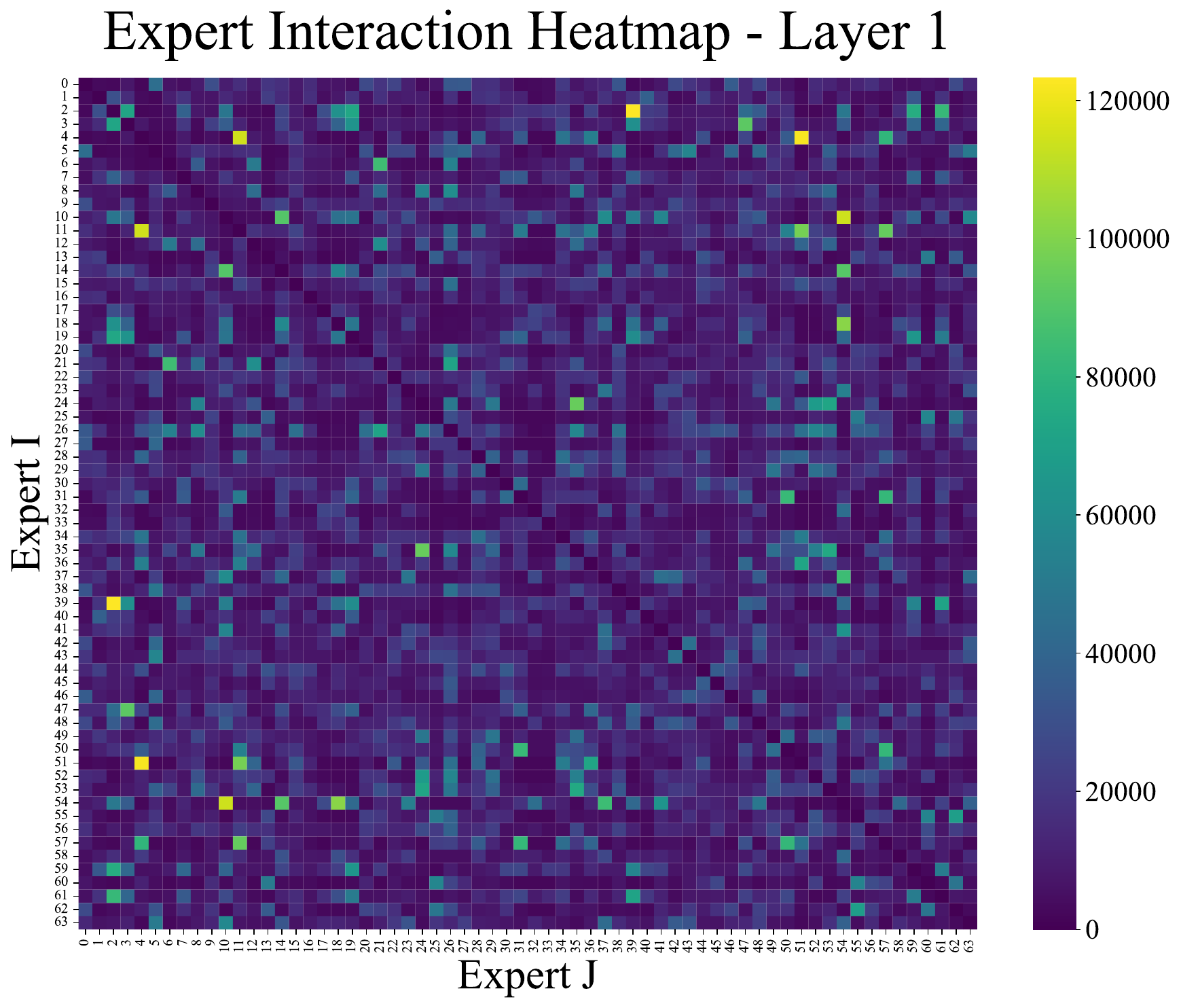}
    \caption{Expert co-activation heatmap for Layer 1. Each cell ($i$, $j$) indicates how frequently experts $i$ and $j$ were selected together for the same token. The sparse, bright pattern demonstrates that co-activation is highly non-uniform: specific pairs of experts are frequently co-activated, suggesting functional redundancy. This pattern forms the basis for identifying "buddy" experts.}
    \label{fig:coactivation_heatmap}
    \vspace{-1em}
\end{figure}

The buddy identification process in BuddyMoE is grounded in two robust empirical regularities observed during model inference: uneven expert activation and sparse co-activation patterns. These properties, visualized in Figure~\ref{fig:expert_activation_distribution} and Figure~\ref{fig:coactivation_heatmap} respectively, reveal that not all experts are equally utilized and that their inter-dependencies are highly structured, which forms the basis of our buddy replacement strategy.

\paragraph{Empirical Evidence from Profiling.}
Over a profiling corpus $\mathcal{D}_\text{prof}$, we record per-layer statistics: (i) per-expert activations $A_\ell(i)$, the number of tokens for which $i\in S_\ell(x)$; and (ii) pairwise co-activations $M_\ell(i,j)$, the number of tokens for which $i,j\in S_\ell(x)$ simultaneously. The conditional co-activation distribution with pivot $i$ is defined as:
\begin{equation}
q_{j\mid i} \;=\; \frac{M_\ell(i,j)}{\sum_{j'} M_\ell(i,j')}, \qquad j\neq i,\quad q_{i\mid i}=0.
\label{eq:qji}
\end{equation}

In practice, $A_\ell(\cdot)$ exhibits a heavy-tailed distribution (Figure~\ref{fig:expert_activation_distribution}) where few "popular" experts dominate activation patterns. More importantly, holding $i$ fixed, the mass of $q_{j\mid i}$ concentrates on a small subset of peers: top-$r$ peers (with $r\!\ll\!E_\ell$) often cover a large majority of the co-activations with $i$. This skew, visible as sparse bright cells in the co-activation matrix (Figure~\ref{fig:coactivation_heatmap}), indicates that a handful of peers systematically appear with $i$, suggesting functional similarity and redundancy.

\paragraph{From Evidence to Buddy Definition.}
For each pivot $i$, we sort peers by $q_{j\mid i}$ to obtain the expert sequence $\pi_i(1),\pi_i(2),\dots$ with $q_{\pi_i(1)\mid i}\ge q_{\pi_i(2)\mid i}\ge\dots$. We call $j$ a buddy of $i$ if it lies within the Cumulative Frequency Threshold (CFT) defined coverage of this sequence. Intuitively, the buddy list is a small, high-coverage set of peers that most frequently co-activate with $i$ and are therefore most suitable substitutes when $i$ is unavailable on the GPU. 

\paragraph{Layer-wise Heterogeneity.}
The co-activation patterns exhibit layer-wise heterogeneity: early layers tend to show broader redundancy with more diffuse co-activation patterns, while later layers are more specialized with tighter expert clusters. This heterogeneity informs our per-layer calibration of the CFT parameter $\alpha_\ell$ and allows us to adapt the buddy list sizes across the model depth.

\subsection{Buddy Replacing Mechanism}
\label{sec:buddy-mechanism}

The buddy replacing mechanism converts co-activation patterns into runtime replacement decisions through a two-stage process.

\paragraph{Offline Profiling and Buddy List Construction.}
The offline stage transforms raw profiling data into compact buddy lists through the Cumulative Frequency Threshold (CFT) mechanism. We collect router traces over a held-out profiling set $\mathcal{D}_\text{prof}$ that matches the intended deployment domain. To stabilize estimates: (i) we accumulate both binary co-activations and probability-weighted co-activations ($\sum_x \mathbb{1}\{i,j\in S_\ell(x)\}\cdot \min(\tilde{p}_\ell(i|x),\tilde{p}_\ell(j|x))$); (ii) we apply Laplace smoothing $M_\ell \!\leftarrow\! M_\ell+\epsilon$ before normalization; (iii) we optionally down-weight early warm-up steps to avoid cold-cache artifacts.

Given $\alpha\in(0,1]$, we define the minimal prefix size:
\begin{equation}
t_i(\alpha) \;=\; \min \Bigl\{ t\ \Big|\ \sum_{r=1}^{t} q_{\pi_i(r)\mid i} \ge \alpha \Bigr\}.
\label{eq:ti}
\end{equation}
The buddy list of $i$ is then:
\begin{equation}
\mathcal{B}_\ell(i;\alpha)\;=\;\{\pi_i(1),\,\pi_i(2),\,\dots,\,\pi_i(t_i(\alpha))\}.
\label{eq:buddyset}
\end{equation}
CFT directly encodes our design goal: a small set that covers most co-activations for $i$. Larger $\alpha$ yields broader coverage (higher chance that some buddy is on the GPU), while smaller $\alpha$ is more conservative (tighter similarity, smaller lists). We cap $|\mathcal{B}_\ell(i;\alpha)|\le K_{\max}$ for metadata control and ensure $t_i(\alpha)\ge 1$ for any $i$ with nonzero activity. We allow a per-layer $\alpha_\ell$ or a monotone schedule, and report $|\mathcal{B}_\ell(i;\alpha_\ell)|$ distributions to verify compactness.

\paragraph{Runtime Buddy Replacement.}
During runtime inference, the buddy replacing mechanism executes for each token and layer when CPU-resident experts are requested. The system first checks whether replacement is permitted through the TAE and distribution gates described in Section~\ref{sec:three-metrics}. If both gates allow replacement, the system searches the precomputed buddy list $\mathcal{B}_\ell(i;\alpha)$ for the missing expert $i$ to find a suitable GPU-resident substitute. The selection prioritizes buddies based on their co-activation frequency $q_{j\mid i}$, with optional adjustments for token-specific compatibility and topology awareness. This runtime mechanism ensures that replacements are made judiciously, maintaining model accuracy while minimizing memory transfer overhead.

\paragraph{Sharding and Topology Awareness.}
In distributed settings with tensor or pipeline parallelism, the buddy mechanism adapts to the hardware topology. If buddies cross partition boundaries, we penalize candidates that require cross-link traffic through the topology penalty term in the selection score $\Psi_\ell$. This ensures that buddy replacements preferentially utilize locally available experts, further reducing communication overhead.

\subsection{System Overview and Integration}
\label{sec:system-overview}

BuddyMoE integrates the co-activation analysis, buddy replacing mechanism, and three key metrics into a cohesive system that achieves high throughput under memory constraints while preserving model accuracy.

\paragraph{End-to-End Workflow.}
The complete BuddyMoE workflow operates in two phases. During the offline phase, we profile the model on representative data to construct the co-activation matrix $M_\ell(i,j)$, derive expert sequences $\pi_i(\cdot)$ for each pivot expert, and apply the Cumulative Frequency Threshold to generate compact buddy lists $\mathcal{B}_\ell(i;\alpha)$. These precomputed structures encode the functional redundancy patterns specific to each model and deployment domain.

During the online inference phase, for each token and layer, the system executes a three-stage decision pipeline. First, the Token Activating Entropy gate assesses whether the token is sensitive to expert substitution based on its routing distribution entropy. Second, the distribution gate evaluates the batch-level CPU residency ratio to prevent broad replacements when many experts are offloaded. Third, if both gates permit, the buddy selection mechanism identifies the best available substitute from the precomputed buddy list, prioritizing based on global co-activation frequency, local token compatibility, and topology considerations.

\paragraph{Design Principles and Trade-offs.}
BuddyMoE's design reflects several key principles. The system prioritizes accuracy preservation through conservative gating: tokens with strong expert preferences are never substituted, and replacements are avoided when too many experts are on CPU. The use of precomputed buddy lists eliminates runtime profiling overhead while capturing model-specific redundancy patterns. The three-metric framework provides multiple safety valves, each addressing a different failure mode: token sensitivity (TAE) prevents quality degradation on important tokens, distribution gating ($\delta$) avoids cascading errors from broad replacement, and buddy selection scoring ($\Psi$) ensures the best available substitute is chosen.

The system parameters ($\tau$, $\beta$, $\alpha$) offer deployment-time trade-offs between throughput and accuracy. Conservative settings (low $\tau$, low $\beta$, low $\alpha$) maximize accuracy at the cost of more memory transfers, while aggressive settings reduce transfers but may impact model quality. The layer-wise calibration capability allows these trade-offs to be tuned per layer based on the observed redundancy patterns, with early layers typically tolerating more aggressive replacement than specialized later layers.

\paragraph{Memory and Computational Efficiency.}
BuddyMoE adds minimal overhead to the baseline MoE inference path. The buddy lists require only $O(K_{\max} \cdot E_\ell)$ storage per layer, where $K_{\max}$ is typically small (5-20). The TAE computation reuses the existing router probabilities with only an entropy calculation added. The distribution gate requires a simple count of CPU-resident experts in the batch. Buddy selection is a lookup in the precomputed list followed by a residency check. These operations are negligible compared to the expert forward pass computations they enable by avoiding memory transfers.

\paragraph{Integration with Existing Systems.}
BuddyMoE integrates seamlessly with existing MoE serving infrastructure. The system preserves the standard MoE interface and requires no changes to model weights or router architecture. It operates as a runtime layer between the router and expert execution, making replacement decisions transparent to both upstream (router) and downstream (expert computation) components. The profiling phase can leverage existing training or validation pipelines, and the buddy lists can be serialized and distributed alongside model checkpoints. For systems with existing prefetching or caching mechanisms, BuddyMoE complements these approaches by reducing the pressure on the memory hierarchy when prefetching misses occur.

\section{Implementation}

To mitigate the significant latency incurred by swapping MoE layers between host and device memory, we propose a dynamic, post-processing mechanism termed \textit{Buddy Expert Substitution}. This technique operates immediately after the gating network selects the \textit{top-k} experts for each token. The core principle is to preemptively replace a selected expert that is not resident in GPU memory with a functionally similar "buddy" expert that is already cached on the device. This substitution avoids high-latency data transfers from the host, thereby maintaining computational throughput. Our approach integrates three key components: a pre-computed expert similarity profile, a substitution algorithm that preserves expert diversity, and a highly parallelized implementation designed for minimal overhead.

The foundation of our method is a static, pre-computed buddy profile that quantifies the functional similarity between all experts within each MoE layer. This profile, generated offline, establishes a ranked list of the most suitable substitutes for every expert in the model. The substitution logic, detailed in Algorithm~\ref{alg:buddy_substitution}, is executed for each token in the batch. For every expert assigned to a token, the algorithm first checks its GPU residency status using the boolean mask $\mathcal{M}$. If an expert is not resident, a search for a substitute is initiated by iterating through its ranked list of buddies in the profile $\mathcal{B}$, up to a search limit $H$. A candidate buddy is deemed suitable only if it is resident on the GPU and is not already part of the token's active expert set, $\mathcal{U}_t$. Upon finding the first suitable buddy, the substitution is performed, the active expert set is updated, and the algorithm proceeds to the next expert for that token.

\begin{algorithm}[t]
\caption{The Buddy Expert Substitution algorithm for a single MoE layer.}
\label{alg:buddy_substitution}
\begin{algorithmic}[1]
\State \textbf{Input:}
\State \quad $\mathcal{S} \in \mathbb{N}^{T \times K}$: Matrix of selected expert indices for $T$ tokens and $K$ experts per token.
\State \quad $\mathcal{M} \in \{\text{true}, \text{false}\}^{E}$: Boolean mask of GPU residency for $E$ experts.
\State \quad $\mathcal{B} \in \mathbb{N}^{E \times R_{\max}}$: Pre-computed buddy profile, where $\mathcal{B}[i, j]$ is the $j$-th best buddy for expert $i$.
\State \quad $H \in \mathbb{N}^+$: Hyperparameter defining the maximum buddy search rank.
\State \textbf{Output:}
\State \quad $\mathcal{S}'$: The modified matrix of expert indices.

\Procedure{BuddySubstitute}{$\mathcal{S}, \mathcal{M}, \mathcal{B}, H$}
    \State $\mathcal{S}' \gets \mathcal{S}$
    \For{each token $t$ from $1$ to $T$}
        \State Let $\mathcal{U}_t$ be the set of unique experts assigned to token $t$ in $\mathcal{S}'$.
        \For{each expert rank $k$ from $1$ to $K$}
            \State $e_{id} \gets \mathcal{S}'[t, k]$
            \If{$\mathcal{M}[e_{id}]$ is false}
                \For{rank $r$ from $1$ to $H$}
                    \State $b_{id} \gets \mathcal{B}[e_{id}, r]$
                    \If{$\mathcal{M}[b_{id}]$ is true \textbf{and} $b_{id} \notin \mathcal{U}_t$}
                        \State $\mathcal{S}'[t, k] \gets b_{id}$
                        \State $\mathcal{U}_t \gets \mathcal{U}_t \cup \{b_{id}\}$
                        \State \textbf{break}
                    \EndIf
                \EndFor
            \EndIf
        \EndFor
    \EndFor
    \State \textbf{return} $\mathcal{S}'$
\EndProcedure
\end{algorithmic}
\end{algorithm}

To ensure this substitution process introduces negligible latency, the entire logic is encapsulated within a custom CUDA kernel. The workload is partitioned across the GPU by assigning a dedicated CUDA thread block to manage the substitutions for a single token, and within each block, a distinct thread is assigned to each of the token's \textit{top-k} experts. This structure enables concurrent evaluation and replacement. A critical challenge in this parallel design is efficiently enforcing the uniqueness constraint for substitutes (i.e., that $b_{id} \notin \mathcal{U}_t$). We resolve this by employing block-level shared memory to maintain a record of the experts already assigned to the token. When a thread identifies a viable buddy, it uses an atomic compare-and-swap operation to claim that buddy expert, a lock-free mechanism that prevents race conditions where multiple threads might otherwise select the same substitute simultaneously. This GPU-native implementation ensures that the substitution logic is executed with high efficiency, effectively decoupling the MoE computation from I/O-bound expert swapping.

\section{Evaluation}

This section presents a comprehensive evaluation of BuddyMoE, a system designed to optimize MoE model inference in memory-constrained environments. We analyze how BuddyMoE maintains model accuracy while improving inference throughput through strategic expert replacement based on co-activation patterns.

\subsection{Experimental Setup}

\noindent\textbf{Model and Hardware Configuration.}
We evaluate BuddyMoE using DeepSeek-V2-Lite, configured with 64 experts per MoE layer and a top-6 gating policy that activates six experts per token. This configuration represents a typical production deployment scenario for large-scale MoE models. Our experimental infrastructure consists of a single-node system equipped with an NVIDIA A100 GPU (PCIe interface) and an Intel Xeon Platinum 8457C processor. We intentionally use a single-GPU configuration with CPU offloading to evaluate performance characteristics relevant to resource-constrained deployment scenarios.

\noindent\textbf{Software Environment and Methodology.}
We implement BuddyMoE within the llama.cpp framework, utilizing its CUDA backend for GPU kernel execution and native CPU offloading mechanisms for expert management. All experiments maintain default llama.cpp configurations for numerical precision, tokenization, and key-value caching to ensure reproducibility. Performance evaluation focuses on two primary metrics: model accuracy, measured using ARC-Easy and ARC-Challenge benchmarks (reported as individual scores and arithmetic mean), and inference throughput, measured in tokens per second including all system overheads from CPU-GPU transfers, routing decisions, and expert prefetching.

\noindent\textbf{System Parameters and Baselines.}
BuddyMoE provides several parameters to control the accuracy-throughput trade-off. The cache rate $c \in \{0.375, 0.50, 0.75\}$ determines the fraction of experts retained in GPU memory. The co-activation frequency threshold (CFT) $\tau$ controls buddy set construction, resulting in buddy set sizes $|\mathcal{B}|$ typically ranging from 2 to 16 experts. The replacement budget $\rho$ limits the maximum number of expert replacements per token, providing fine-grained control over the replacement strategy.

We compare BuddyMoE against two baseline approaches. The original baseline uses only true experts without replacement, representing maximum accuracy but potentially incurring prefetch-induced latency penalties. The random replacement baseline substitutes missing experts with uniformly selected GPU-resident experts, providing a naive comparison point for our co-activation-based strategy.

\subsection{Performance Analysis}

\noindent\textbf{High Cache Rate Performance (c = 0.75): }Table~\ref{tab:abl-c075} presents performance characteristics when 75\% of experts remain GPU-resident, representing a moderately constrained scenario. The original baseline achieves maximum accuracy of 0.735 with throughput of 34.23 tokens per second. Random replacement improves throughput to 39.67 t/s (Token Per Second) but severely degrades accuracy to 0.55, demonstrating the inadequacy of uninformed replacement strategies.

\begin{table}[t]
  \centering
  \footnotesize
  \caption{Performance at cache rate $c = 0.75$}
  \label{tab:abl-c075}
  \begin{tabular}{@{}lcccccc@{}}
    \toprule
    \textbf{Method} & \textbf{$\tau$} & \textbf{$|\mathcal{B}|$} & \textbf{$\rho$} & \textbf{ARC-E} & \textbf{ARC-C} & \textbf{Avg / t/s} \\
    \midrule
    Original & -- & -- & -- & 0.81 & 0.66 & 0.735 / 34.23 \\
    Random & -- & -- & -- & 0.62 & 0.48 & 0.55 / 39.67 \\
    \midrule
    BuddyMoE & 0.75 & 4 & -- & 0.70 & 0.58 & 0.64 / 37.42 \\
    BuddyMoE & 0.95 & 16 & -- & 0.62 & 0.43 & 0.525 / 37.12 \\
    BuddyMoE & 0.95 & 16 & 3 & 0.76 & 0.63 & \textbf{0.695} / 36.75 \\
    BuddyMoE & 0.95 & 16 & 4 & 0.60 & 0.48 & 0.54 / 38.50 \\
    \bottomrule
  \end{tabular}
\end{table}

BuddyMoE demonstrates superior performance across multiple configurations. The optimal configuration ($\tau = 0.95$, $|\mathcal{B}| = 16$, $\rho = 3$) achieves 0.695 average accuracy at 36.75 t/s. This represents only a 5.4\% accuracy reduction compared to the original baseline while delivering 7.4\% higher throughput. Importantly, this configuration maintains 26.4\% higher accuracy than random replacement while achieving comparable throughput, validating our co-activation-based approach.

\noindent\textbf{Moderate Cache Rate Performance (c = 0.50):}
Table~\ref{tab:abl-c050} examines system behavior under increased memory pressure, with only 50\% of experts retained in GPU memory. Random replacement catastrophically fails under these conditions, achieving only 0.23 average accuracy despite maintaining 33.14 t/s throughput, rendering the model effectively unusable.

\begin{table}[t]
  \centering
  \footnotesize
  \caption{Performance at cache rate $c = 0.50$}
  \label{tab:abl-c050}
  \begin{tabular}{@{}lcccccc@{}}
    \toprule
    \textbf{Method} & \textbf{$\tau$} & \textbf{$|\mathcal{B}|$} & \textbf{$\rho$} & \textbf{ARC-E} & \textbf{ARC-C} & \textbf{Avg / t/s} \\
    \midrule
    Original & -- & -- & -- & -- & -- & -- / 28.56 \\
    Random & -- & -- & -- & 0.28 & 0.18 & 0.23 / 33.14 \\
    \midrule
    BuddyMoE & 0.99 & 2 & -- & 0.62 & 0.44 & 0.53 / 28.95 \\
    BuddyMoE & 0.95 & 16 & -- & 0.36 & 0.24 & 0.30 / 31.87 \\
    BuddyMoE & 0.95 & 16 & 3 & 0.72 & 0.55 & \textbf{0.635} / 30.21 \\
    BuddyMoE & 0.95 & 16 & 4 & 0.58 & 0.52 & 0.55 / 31.02 \\
    \bottomrule
  \end{tabular}
\end{table}

BuddyMoE demonstrates good resilience under these constraints. The configuration with replacement budget $\rho = 3$ achieves 0.635 average accuracy at 30.21 t/s, representing a 176\% improvement over random replacement while maintaining competitive throughput. Smaller buddy sets ($|\mathcal{B}| = 2$) with stricter CFT thresholds ($\tau = 0.99$) maintain reasonable accuracy (0.53) but incur modest throughput penalties, illustrating the trade-offs between buddy set size and replacement effectiveness.

\noindent\textbf{Extreme Memory Constraint Performance (c = 0.375):}
Table~\ref{tab:abl-c0375} evaluates performance under extreme memory constraints, with only 37.5\% of experts GPU-resident. This scenario represents the practical limit of memory-constrained deployment. The original baseline achieves 24.78 t/s under these conditions, while random replacement produces unusable accuracy of 0.16.

\begin{table}[t]
  \centering
  \footnotesize
  \caption{Performance at cache rate $c = 0.375$}
  \label{tab:abl-c0375}
  \begin{tabular}{@{}lcccccc@{}}
    \toprule
    \textbf{Method} & \textbf{$\tau$} & \textbf{$|\mathcal{B}|$} & \textbf{$\rho$} & \textbf{ARC-E} & \textbf{ARC-C} & \textbf{Avg / t/s} \\
    \midrule
    Original & -- & -- & -- & -- & -- & -- / 24.78 \\
    Random & -- & -- & -- & 0.14 & 0.18 & 0.16 / -- \\
    \midrule
    BuddyMoE & 0.95 & 16 & -- & 0.21 & 0.14 & 0.175 / 30.22 \\
    BuddyMoE & 0.95 & 16 & 3 & 0.76 & 0.53 & \textbf{0.645} / 27.33 \\
    BuddyMoE & 0.95 & 16 & 4 & 0.51 & 0.51 & 0.51 / 27.46 \\
    \bottomrule
  \end{tabular}
\end{table}

BuddyMoE with $\rho = 3$ maintains functional performance with 0.645 average accuracy at 27.33 t/s, achieving 10.3\% higher throughput than the original baseline. This remarkable accuracy retention under severe memory constraints demonstrates the effectiveness of co-activation-based expert replacement. The system maintains usable model performance in scenarios where conventional approaches fail entirely.

\subsection{PCIe Bandwidth Usage}

\begin{figure}
    \centering
    \includegraphics[width=0.9\linewidth]{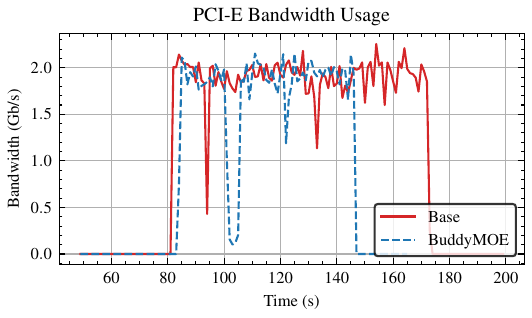}
    \caption{A comparison of PCIe bandwidth trends for the Base and Buddy-MoE methods. The chart highlights the efficiency of the Buddy-MoE method, which uses 20\% less bandwidth.}
    \label{fig:pcie-bw}
\end{figure}

We tested the PCIe bandwidth for different methods of handling data prefetch failures. Our test measured how much data was sent over the PCIe bus. As shown in Figure~\ref{fig:pcie-bw} a clear difference: the Base method used about 20\% more PCIe bandwidth than the \texttt{BuddyMoE} method for PCIe reads. This is because the base method transfers missing data from the main computer memory, while the \texttt{BuddyMoE} method only looks for it in the GPU’s own memory.

This difference in how they work shows a key advantage of the \texttt{BuddyMoE} method. By avoiding extra trips to the main memory, it creates less traffic and keeps the bandwidth use more stable. In short, for tasks that need to be fast and efficient, the buddymoe method is a better choice because it puts less strain on the system's main data bus.

\subsection{Analysis of Expert Co-activation Structure}

Figure~\ref{fig:coactivation_heatmap} visualizes the expert co-activation frequency matrix for Layer 1, revealing the structural patterns that enable effective buddy expert identification. The heatmap demonstrates highly non-uniform co-activation frequencies across expert pairs, with specific combinations showing significantly elevated joint activation rates.

\begin{figure*}[h]
    \centering
    \includegraphics[width=1.0\linewidth]{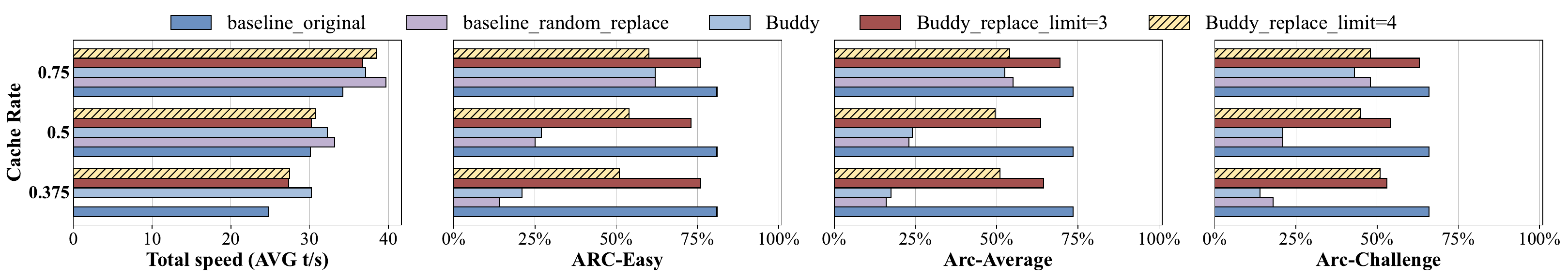}
    \caption{Expert co-activation frequency heatmap for Layer 1. Cell intensities represent the frequency with which expert pairs are jointly activated during token processing. The sparse distribution of high-intensity regions indicates strong functional relationships between specific expert pairs, providing the statistical foundation for buddy expert selection.}
    \label{fig:coactivation_heatmap}
\end{figure*}

These sparse, high-intensity patterns reveal strong functional relationships between specific expert pairs, validating our core hypothesis that certain experts exhibit sufficient functional similarity to serve as effective substitutes during inference. The non-uniform distribution of co-activation frequencies provides the statistical foundation for informed buddy selection, enabling replacement decisions that preserve model functionality while improving throughput. This structural analysis confirms that expert co-activation patterns encode meaningful functional relationships that can be exploited for efficient inference under memory constraints.

\section{Related Work}

\noindent\textbf{Mixture-of-Experts Models.} The MoE architecture scales language models by activating only a sparse subset of parameters per token. Shazeer et al.\ introduced the sparsely-gated MoE layer, where a gating network routes each input to a small subset of expert feed-forward networks instead of a dense layer~\cite{shazeer2017outrageously}. Subsequent large-scale systems, such as Google’s GShard~\cite{lepikhin2021gshard} and the Switch Transformer~\cite{fedus2021switch}, demonstrated the effectiveness of this idea at unprecedented scales. GShard combined conditional computation with automatic sharding and \emph{demonstrated} MoE models beyond 600B parameters (and discussed scaling up to trillions). The Switch Transformer simplified gating to top-1 (a single expert per token) and achieved state-of-the-art results with up to $\sim$1.6T parameters; a representative configuration (Switch-C) uses 2048 experts per layer across 15 layers—i.e., tens of thousands of experts across the network. These works showed MoEs can match or exceed dense models’ quality while activating only a fraction of parameters, setting the stage for modern MoE LLMs (e.g., Mixtral, DeepSeek-MoE, Snowflake Arctic). However, the memory footprint of storing all experts remains a challenge at inference time for models with tens to hundreds of billions of total parameters.

\noindent\textbf{Memory Offloading for Large Models.}
Memory offloading is a common technique to deploy large-scale models on hardware with constrained GPU memory. General-purpose systems like DeepSpeed's ZeRO-Infinity~\cite{rajbhandari2021zero} and Hugging Face Accelerate~\cite{hf_accelerate_infer_2025,hf_accelerate_deepspeed_2025} offload coarse-grained model components such as entire layers to CPU memory or NVMe storage. MoE models, however, present a unique offloading challenge due to their fine-grained, dynamic expert activations. Systems must swap individual experts with low latency, making I/O management critical.

Research in this area has focused on two primary strategies: predictive prefetching and intelligent cache management.
For prefetching, MoE-Infinity~\cite{xue2024moe} traces historical expert usage to predict and load upcoming experts, aiming to hide data transfer latency. Similarly, Pre-gated MoE~\cite{hwang2024pregatedmoe} integrates an auxiliary \emph{pre-gate} into the model architecture to predict the next layer’s expert needs one step ahead.
For cache management, EdgeMoE~\cite{yi2023edgemoe} improves upon standard LRU/LFU eviction policies by designing a heuristic that considers both activation frequency and layer index. It also reduces data transfer volume by applying expert-wise mixed-precision quantization. These works demonstrate that specialized, model-aware strategies are essential for minimizing I/O stalls in offloaded MoE inference.

\smallskip\noindent\textbf{Reducing Expert Loading Latency.} Because on-demand expert loads can dominate end-to-end latency, several strategies aim to mitigate this bottleneck. \emph{Quantization and skipping.} EdgeMoE uses expert-wise mixed-precision (e.g., assigning lower bit-width to less critical experts) to reduce memory and transfer cost with limited impact on accuracy~\cite{yi2023edgemoe}. AdapMoE adaptively limits the number of activated experts $k$ based on a sensitivity metric, integrating improved prefetching and caching; it reduces the average number of experts by $\sim$25\% and reports $\sim$1.35$\times$ speedup without accuracy loss on edge platforms~\cite{zhong2024adapmoe}. \emph{Dynamic expert pools.} SwapMoE maintains a small set of high-value “virtual experts’’ in GPU memory and swaps others to CPU, showing reduced memory consumption (e.g., 14.2$\rightarrow$4.7\,GiB) and $\sim$50\% latency reduction on Switch Transformer benchmarks, while preserving quality~\cite{kong2024swapmoe}. Complementary to these, Fate uses cross-layer gate signals to prefetch next-layer experts and a shallow-favoring cache, reporting expert-hit rates $\sim$99\% and up to $4.1\times$ decoding speedups under offloading~\cite{fang2025fate}. In parallel, Lu et al.\ study post-training expert pruning/skipping, finding that removing or skipping infrequently used experts often yields negligible loss on downstream tasks, reinforcing redundancy in large MoEs~\cite{lu2024experts}. Overall, these methods leverage redundancy and variable expert importance to reduce effective model size or avoid expensive loads, trading negligible accuracy for large efficiency gains.


\section{Conclusion}

The rapid scaling of MoE models has created a deep tension between their computational sparsity and their memory density. Conventional offloading and prefetching systems treat this as a pure I/O optimization problem, striving for perfect prediction in an inherently unpredictable environment. This work proposes a paradigm shift. With BuddyMoE, we argue that the redundancy in over-parameterized models is not just an inefficiency to be pruned, but a resource to be exploited for system resilience.

By substituting functionally equivalent experts at runtime, we reframe the costly penalty of a prefetch miss as a low-cost, on-the-fly model approximation. This principle of "runtime approximation for system performance" opens new avenues for co-designing LLM architectures and inference systems. Our results show that this approach is highly effective, drastically reducing latency with negligible accuracy loss. As models continue to outgrow hardware, embracing such controlled trade-offs will be essential for making their power accessible and efficient.



\bibliography{bibfile}

\end{document}